\documentclass[accepted]{article}

\usepackage{microtype}
\usepackage{graphicx}
\usepackage{booktabs} 
\usepackage{amssymb}
\usepackage{amsmath}
\usepackage{subcaption}
\usepackage{graphicx}
\usepackage{soul,color}
\usepackage{multirow}
\usepackage{bm} 
\usepackage{balance}

\usepackage{hyperref}

\newcommand{\addReviewer}[2]{
  \expandafter\newcommand\csname #1\endcsname[1]{{\bf \color{#2} \expandafter\MakeUppercase #1:\,##1}}
}

\definecolor{asparagus}{rgb}{0.53, 0.66, 0.42}
\definecolor{alizarin}{rgb}{0.82, 0.1, 0.26}
\addReviewer{jary}{asparagus}
\addReviewer{simone}{alizarin}

\usepackage{icml_style/icml2020}

\icmltitlerunning{Pseudo-Rehearsal for Continual Learning with Normalizing Flows}

\begin{document}

\twocolumn[
\icmltitle{Pseudo-Rehearsal for Continual Learning with Normalizing Flows}



\icmlsetsymbol{equal}{*}

\begin{icmlauthorlist}
\icmlauthor{Jary Pomponi}{equal,diet}
\icmlauthor{Simone Scardapane}{equal,diet}
\icmlauthor{Aurelio Uncini}{diet}

\end{icmlauthorlist}

\icmlaffiliation{diet}{Department of Information Engineering, Electronics and Telecommunications (DIET), Sapienza University of Rome, Italy}

\icmlcorrespondingauthor{Jary Pomponi}{jary.pomponi@uniroma1.it}

\icmlkeywords{Machine Learning, Continuous Learning, Normalizing Flow, catastrophic Forgetting}

\vskip 0.3in
]



\printAffiliationsAndNotice{\icmlEqualContribution} 

\begin{abstract}
Catastrophic forgetting (CF) happens whenever a neural network overwrites past knowledge while being trained on new tasks. 
Common techniques to handle CF include regularization of the weights (using, e.g., their importance on past tasks), and rehearsal strategies, where the network is constantly re-trained on past data. Generative models have also been applied for the latter, in order to have endless sources of data. In this paper, we propose a novel method that combines the strengths of regularization and generative-based rehearsal approaches. Our generative model consists of a normalizing flow (NF), a probabilistic and invertible neural network, trained on the internal embeddings of the network. By keeping a single NF conditioned on the task, we show that our memory overhead remains constant. In addition, exploiting the invertibility of the NF, we propose a simple approach to regularize the network's embeddings with respect to past tasks. We show that our method performs favorably with respect to state-of-the-art approaches in the literature, with bounded computational power and memory overheads.
\end{abstract}

\section{Introduction}
\label{section:introduction}


\begin{figure}[t!]
\vskip 0.1in
\begin{center}
\begin{subfigure}[h]{0.48\textwidth}
\includegraphics[width=3in]{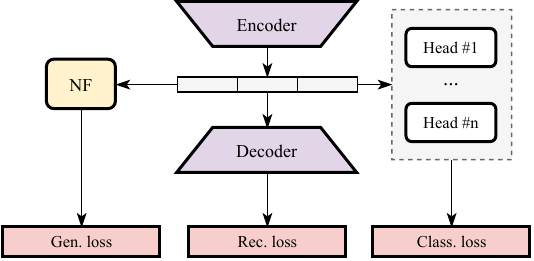}
\caption{Training step}
\label{fig:overview}
\end{subfigure}
\vskip 0.1in
\begin{subfigure}[h]{0.48\textwidth}
\includegraphics[width=3.1in]{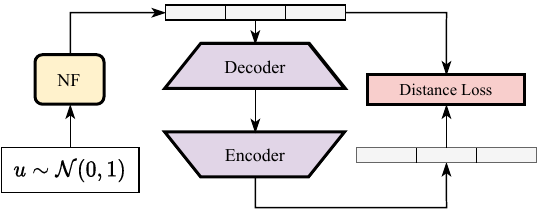}
\caption{Regularization step}
\label{fig:training}
\end{subfigure}%
\caption{Proposed algorithm for continual learning. (a) Training step, in which both the classifier network (encoder with the current head) and the generative model (NF with the decoder) are trained: the first one to solve the task, and the second one to replicate the samples from all the past tasks. (b) Regularization step, in which the encoder is regularized so that the new extracted embeddings are as close as possible to the past ones.}
\end{center}
\vskip -0.2in
\end{figure}

One of the major open problems in deep learning is the so-called catastrophic forgetting (CF) \cite{MCCLOSKEY1989109, Ratcliff1990ConnectionistMO, french1999catastrophic}. It is the tendency of a NN to forget past learned information when training on new tasks. This problem is intrinsically connected with the continual learning (CL) property of a NN, which is the ability of a NN to learn consecutive information without forgetting previously stored knowledge.    



Overcoming, or mitigating, CF is a key step in order to achieve a more general artificial intelligence; a system should be able to learn a sequence of tasks and remember them, following the lifelong learning paradigm \cite{THRUN199525}. This is a key problem because real world tasks continually evolve and, often, it is not possible to train a NN from scratch. Without efficient methods to overcome CF, online training in a lifelong learning scenario is not possible. Recently, there has been a resurgence of interest in this area due to its importance.

One of the first attempts to mitigate CF consisted in storing past examples and replaying them into the model while learning new information \cite{robins1995catastrophic}. These \textit{rehearsal} methods have been improved over the years, with more complex memory systems and hybrid approaches emerging. In particular, many \textit{pseudo-rehearsal} methods have been proposed, in which the external memory is replaced with a generative model, capable of generating endless samples from the past. Pseudo-rehearsal algorithms, however, require generative models that are both easy to train (in order to provide high-quality data) and simple to condition on the task (in order to avoid having one generative model for each task). Both of these conditions are extremely challenging in practice.

To overcome (or complement) these limitations, many \textit{regularization} methods have been studied over the last years, in which additional loss terms are used to mitigate CF. In elastic weight consolidation (EWC, \cite{Kirkpatrick2017overcoming}), for example, past weights are used to regularize the training process, by slowing down the modification of important weights, where the importance is quantifed based on a Fisher information criterion. Alternative regularization strategies are achieved by acting on the previous outputs \cite{li2017learning} or gradients \cite{lopez2017gradient}. Recently, it was also shown that state-of-the-art performance can be achieved by acting on the internal embeddings (i.e., the activations before the classification layer) \cite{pomponi2020er}. A more complete overview of CL methods is provided later on in Section \ref{section:methods}, or in \cite{parisi2019continual}.


In this paper, we propose a novel CL method aiming to combine the benefits of pseudo-rehearsal and regularization strategies. Our algorithm is shown in Fig. \ref{fig:overview}, and can be summarized in three steps:

\begin{enumerate}
    \item Similarly to pseudo-rehearsal strategies, we store information on past tasks by training an auxiliary generative model. Instead of training it on the input space, however, we train it to generate samples from the internal embeddings of the network (the output of the last convolution layer), simultaneously with the main classifier.
    \item We use a normalizing flow (NF, \cite{papamakarios2019normalizing}) as generative model. NFs are invertible neural networks that can perform sampling and density estimation in both directions. In this way, the NF can be trained efficiently, with no need for additional components such as in generative adversarial networks.
    \item Finally, we use the sampled embeddings from the trained NF to perform regularization with respect to the past tasks, as shown in Fig. \ref{fig:training}.
\end{enumerate}

We claim that (i) training the generative model in the embeddings' space is significantly easier (both in its design and in its optimization) compared to the input space, and (ii) that regularization makes better use of past information stored in the generative model, compared to simply augmenting the mini-batch with new data (similarly to \cite{pomponi2020er}). In our experimental evaluation, we validate these two claims, and we show that our method performs favorably (or better) than several state-of-the-art approaches, while requiring significantly less memory and computational overhead.

\section{Related Works}
\label{section:methods}

The methods for overcoming CF can be categorized, in line with \cite{parisi2019continual} and \cite{maltoni2019continuous}, in three broad groups. We underline that the boundaries are not always defined, with many methods, including ours, exploiting two or more of these strategies.

\begin{itemize}
    \item \textbf{Architectural Strategies}: methods that use specific architectures, layers, activation functions and/or weights freezing/pruning, and eventually grow the architecture when needed (e.g., \cite{rusu2016progressive}). For example, Hard Attention on Task (HAT, \cite{serra2018overcoming}) uses an attention mechanism in order to route the information/gradient flow in the NN and preserve the weights associated to past tasks.
    \item \textbf{Rehearsal strategies}: in this case, past examples are stored and later replayed in the current mini-batches to consolidate the network. In order to avoid having to explicitly store past examples, which requires a growing memory, \textit{pseudo-rehearsal} algorithms \cite{robins1995catastrophic} craft them on-the-fly, most notably exploiting generative models \cite{shin2017continual}.
    \item \textbf{Regularization techniques}: in this case, popularized by elastic weight consolidation (EWC, \cite{Kirkpatrick2017overcoming}), the loss function on the current task is extended with a regularization penalty to selectively consolidate past information or slow the training on new tasks. Broadly speaking, regularization methods are easy to implement, but they require to carefully select what information is being regularized, and how.
\end{itemize}

The method we propose in this paper is at the boundary of pseudo-rehearsal and regularization strategies, so we focus on these two classes below. In fact, our method use a generative model in the embedding space and then use the generated embeddings to regularize the network while training on new tasks.  

Learning Without Forgetting (LWF, \cite{li2017learning}) is one of the first regularization methods. It attempts to alleviate CF by stabilizing the output layer using knowledge distillation. Others well-known regularization methods are EWC, which applies a soft structural regularization computed between the weights' importance relative to the past tasks and the current weights, and Synaptic Intelligence (SI, \cite{zenke2017continual}), a modification of EWC, which uses the difference between the current weights and their trajectory calculated during the training. Other methods include Gradient Episodic Memory (GEM, \cite{lopez2017gradient}), Averaged-GEM \cite{chaudhry2018efficient}, and the recently proposed Embedding Regularization (ER, \cite{pomponi2020er}). 

In GEM, the external memory is populated with past examples that are used to regularize the direction of the current gradients, in order to move the weights in a region of the space in which all the tasks are satisfied. This method is capable of improving past scores, but it requires to solve a complex minimization problem at every step, which does not scale well with the number of tasks. ER is a regularization technique in which the external memory contains past examples and their associated embeddings, extracted at the end of the training process on the associated task. The memory is used to impose a penalty to constrain the current embeddings to lie in the vicinity of the past ones. This method is extremely fast and requires little memory. The approach we propose in this paper follows the philosophy of ER to act at the level of the embeddings, instead of single weights or outputs. Another novel and interesting approach is proposed in \cite{ebrahimi2020adversarial}, in which the authors use an Adversarial Continual Learning (ACL) approach: it aims to alleviate CF by learning a disjoint latent space representation composed of a task-specific latent space for each task and a task-invariant feature space for all tasks.

A more challenging set of methods are pseudo-rehearsal ones. In \cite{shin2017continual}, the authors proposed a method which consists of two modules: a deep generative model and a task solver. In this way, samples from past tasks can be generated using the generative model and interleaved with information from the new tasks; the solver is used to predict the label associated to the generated images and then regularize the network. In \cite{kang2020discriminative} a similar approach, but based on a Variational Autoencoder (VAE), is proposed: it consists in a VAE and an external NN, which learns to replicate the distribution of the embeddings associated to a task; this external NN can be used to generate images associated to past tasks and reduce CF. Right now, pseudo-rehearsal methods were strictly evaluated on datasets of relatively low complexity. The question on whether these generative approaches can scale up to more complex domains is still open. 

The method we propose here can be considered a pseudo-rehearsal one, but we focus on a more recent class of generative models \cite{kingma2018glow}, and we apply them at the level of embeddings instead of in the input space. We note that in the literature on generative models, a number of authors have considered similar combinations of autoencoders with a generative model on their latent space. In \cite{rezende2015variational}, the authors applied a NF to learn a VAE prior. This idea has been further studied in \cite{kingma2016improved}, where the authors proposed a new NF which scales well for high-dimensional embedding spaces. More similar to our proposal, in \cite{guo2019auto} the authors proposed a model which uses an adversarial generative model in the embedding space to generate high resolution images.

A relatively new and emerging area of study researches how to mitigate CF when the tasks' boundaries are not known. In this area we highlight \cite{aljundi2019task}, in which the authors proposed a task free approach to continual learning, using a regularization-based memory. In \cite{zeno2018task}, a task agnostic Bayesian method was proposed, demonstrating the ability of probabilistic models to handle ambiguous task boundaries. Finally, \cite{rao2019continual} has introduced the novel idea of unsupervised learning in a lifelong scenario.


Many other methods exists; for a complete review of existing methods see \cite{parisi2019continual}. 

\section{Proposed method}
\label{section:proposed}

\subsection{Motivation}
\label{sec:motivation}
In a sense, rehearsal methods are close to optimal, because in the limit of a very large memory they recover a standard multi-task setting. On the other hand, these methods require a memory that, usually, grows linearly with the number of tasks. Additionally, the computational requirements increase with the number of tasks, since each past task needs to be remembered (and rehearsed) separately.

Pseudo-rehearsal methods try to overcome these limitations by substituting the memory with a generative model and doing parameter sharing on the generative model, which is incrementally trained on all the tasks and constrained to remember the information about the tasks encountered so far. However, this creates a new set of challenges: (i) the CF problem is removed from the NN, but the generative model itself potentially suffers CF; (ii) doing parameter sharing on real-world images can be difficult. 

The aim of this paper is to propose a generative approach which does not work directly on the input, and that can be used to regularize the model instead of simply augmenting the dataset. The key idea of the proposed pseudo-rehearsal embedding regularization (PRER) is to use a generative model to sample new embeddings associated to past tasks. To regularize the network, we (i) reconstruct the images associated to the embeddings, (ii) calculate the embeddings given by the current network; (iii) force the old and the new embeddings to be as close as possible (by moving the current one in the direction of the past ones). 


\subsection{Problem formulation}
\label{section:formulation}
The following formulation of the CL scenario is similar to the one proposed in \citep{lopez2017gradient}. We receive a sequence of tasks $t=1, \dots, M$, each one composed by a set of triples $\{(\mathbf{x}_i, t, y_i)\}_{i=1}^S \in \mathcal{X} \times \mathbb{N^+} \times \mathcal{Y}$, where $\mathbf{x}_i$ is a sample, $t$ is an integer identifying the current task, and $y_i$ is the corresponding label of $\mathbf{x}_i$. Within this formulation, the tasks never intersect, and a new task is collected only when the current one is over. The input and the labels can belong to any domain, although most benchmarks in the CL literature have considered the image domain in a classification setting \cite{parisi2019continual}.

In this paper we focus on a CL scenario in which each task has its own classifier, meaning that only a portion of the NN, that we call the encoder, is shared (top violet part in Fig. \ref{fig:overview}). There is no class overlap among different tasks, and accuracy is computed separately for each task. This is done by creating a leaf branch for each task, called head, which classifies only images associated to that task (right part in Fig. \ref{fig:overview}). Mathematically, we have a model $f(\mathbf{x}, t) = S_t(E(\mathbf{x}))$, where $E(\cdot)$ is the encoder and $S_t(\cdot)$ a task-specific classifier. In a naive setting, every time we receive a new task $t$, we minimize a task-specific loss starting from the current encoder $E$ and the new randomly-initialized $S_t$:
\begin{equation}
\mathcal{L}_t = \sum_{i=1}^S L(f(\mathbf{x}_i, t), y_i) \,,
\label{eq:classifier_loss}
\end{equation}
where $L$ is a suitable loss (e.g., cross-entropy). CF appears whenever training on the current task $t$ degrades the performances on previous tasks $1, \ldots, t-1$.

Such models cannot be used to classify unknown samples, since the belonging task needs to be known a priori. These experiments are suitable for studying the feasibility of training disjoints tasks without forgetting how to solve the previous ones. We leave explorations of alternative scenarios, such as the so-called Single Incremental Task (SIT) \cite{maltoni2019continuous}, for future work.

\subsection{Pseudo-rehearsal in the embedding space}
As shown in Fig. \ref{fig:overview}, we augment the classifier described in Section \ref{section:formulation} with two additional networks: 
\begin{enumerate}
    \item Firstly, we add a decoder network $\hat{\mathbf{x}} = D(\mathbf{z})$ to approximately invert the encoder network. Both of the networks are trained jointly as an autoencoder network. 
    \item Secondly, we train a generative model to generate samples from $p(\mathbf{z} \,\vert\, t, y)$, i.e., embeddings of a specific task and class. This model is implemented with a NF, described in Section \ref{section:nf}.
\end{enumerate}

Given the data for the new task, the training process is divided in three phases:

\textbf{1) Autoencoder training}: In the first stage we train the encoder and decoder networks on the current task, by minimizing a reconstruction loss:
\begin{equation}
    \mathcal{L}_{ae} = \sum_{i=1}^S \lVert \mathbf{x}_i - D(E(\mathbf{x}_i)) \rVert^2 \,.
    \label{eq:autoencoder_loss}
\end{equation}

To avoid to use all the capacity of the autoecnoder we implement it as a sparse autoencoder \cite{arpit2015regularized}. In this way, by regularizing the layers of our network, not only the reconstruction improves but also the classification scores, due to the sparsity of the produced embeddings.

\textbf{1.1) Pseudo-rehearsal embedding regularization}

If the task is not the first one, we further overwrite a percentage of each mini-batch with samples drawn from the generative model. We sample from the generative model using the same proportions of labels observed in the training sets. 
In addition to using the generative model to augment the dataset, we also apply a regularization technique similar to the ER method proposed in \cite{pomponi2020er}. This step is applied only while we train the autoencoder and only its weights are updated; this step is is depicted in Fig. \ref{fig:training}.

Consider a generic sample $\hat{\mathbf{z}} \sim p(\mathbf{z} \,\vert\, t, y)$ from the generative model. We first project the embedding to the input space using the trained decoder $\hat{\mathbf{x}} = D(\hat{\mathbf{z}})$. Then, we regularize the autoencoder by penalizing deviations from the currently generated embeddings:
\begin{equation}
    R(\hat{\mathbf{z}}) = d(\hat{\mathbf{z}}, E(D(\hat{\mathbf{z}})))\,,
    \label{eq:embedding_regulatization}
\end{equation}
where $d$ is a distance function. Practically, when training the autoencoder, we augment the reconstruction loss in \eqref{eq:autoencoder_loss} by computing \eqref{eq:embedding_regulatization} on all the data we use to augment the dataset.

\textbf{2) Generative model training}: In the second stage, we train the generative model to accurately sample embeddings from the new task $t$. This procedure is described in Section \ref{section:nf}. We underline that we use a single generative model conditioned on the task, in order to keep the memory overhead of our method constant. As in the autoencoder training, if the task is not the first one, we overwrite a percentage of each mini-batch with samples associated from the past tasks, drawn from the generative model. In this phase, regularize the embeddings is not needed, since the NF is will be trained on new and past embeddings jointly; the training itself acts like a consolidation of past information. 

\textbf{3) Classifier training}: In the third stage, the current task-specific head $S_t$ is trained to solve the current task $t$ by minimizing \eqref{eq:classifier_loss}.

We found that splitting the training into three separate processes helps the stability and improves the results, since each block works with the others already optimized. The computational overhead is also small, since each step updates a small part of the overall architecture (we provide memory and time comparisons in our experimental evaluation).

\subsection{Normalizing Flows}
\label{section:nf}

To complete the specification of our PRER method, we need to describe a specific generative model to estimate samples from the embedding space. While most generative methods could be used here, we have found normalizing flows (NFs) to be particularly effective for the task. Introduced in \cite{Tabak2010Density} and \cite{Tabak2013Nonparametric}, and popularized by \cite{rezende2015variational} and \cite{dinh2014nice}, NFs are probabilistic models describing the transformation of a probability distribution into a more complex one using a sequence of differentiable, invertible mapping functions. Depending on the type, a NF is capable of efficiently performing sampling or density estimation in either direction \cite{papamakarios2019normalizing}.

Ignoring for a moment the conditioning on a given task, let $\mathbf{u} \sim p_{u}(\mathbf{u})$ be a real $d$-dimensional distribution which is easy to sample from, and $T$ a transformation $T : \mathbb{R}^d \rightarrow \mathbb{R}^d$. The key requirements for $T$ to be a NF are: 1) $T$ must be invertible, with an inverse denoted by $T^{-1}$, and 2) both $T$ and $T^{-1}$ must be differentiable. We can obtain samples from $p_z(\mathbf{z})$ by drawing samples from the easier $p_u(\mathbf{u})$ and then computing $T(\mathbf{u})$ (forward mapping). Alternatively, we can `normalize' a known sample $\mathbf{z} \sim p_z(\mathbf{z})$ by applying the inverse transformation $T^{-1}(\mathbf{z})$ (inverse mapping).\footnote{As mentioned in \cite{papamakarios2019normalizing}, the terms ``forward'' and ``inverse'' are simply a convention.} Additionally, we can evaluate the likelihood of a known sample $\mathbf{z}$ as:
\begin{align}
    p_{z}(\mathbf{z}) = p_{u}(\mathbf{u}) \Big\vert \text{det}\ J_{T}(T^{-1}(\mathbf{x})) \Big\vert ^ {-1} \,,
    \label{eq:change_of_variable}
\end{align}
where $J_{T^{-1}} \in \mathbb{R}^{d \times d} $ is the Jacobian matrix of all partial derivatives of $T^{-1}$. The prior distribution $p_{u}(\mathbf{u})$ is generally chosen as an Isotropic Normal distribution. $T$ is instead obtained by composing multiple simpler, invertible transformations $T_1$, $\ldots$, $T_L$, resulting in $T = T_L \circ T_{L-1} \circ \dots \circ T_1$. The arbitrarily complex density $\mathbf{z}$ can be constructed from a prior distribution by composing several simple maps and then applying Eq. \ref{eq:change_of_variable}.



As stated before, a NF can perform both sampling and density estimation in both directions. Depending on the specific NF, not all of these operations are necessarily easy to compute \cite{papamakarios2019normalizing}. This is due to the fact that to achieve them simultaneously, the model needs both a simple forward mapping, a simple inverse mapping, and the Jacobians have to be easy to compute. In general, this is not always possible, mostly due to a computational trade-off. Hence, a NF model needs to be built depending on the application.

In our case, we need efficient density estimation and sampling in the inverse mapping (for training), but only efficient sampling in the forward direction (see Fig. \ref{fig:training}). We use a setup similar to \cite{dinh2016density,kingma2018glow}, whose layers we briefly summarize below. Denote by $\mathbf{u}_{j+1} = T_j(\mathbf{u})$ the generic $j$th block of the NF. We build the overall NF by interleaving three types of invertible transformations.

\textbf{1) Coupling Layer and Affine transformation}: Proposed in \cite{dinh2014nice, dinh2016density}, a coupling layer consists in a powerful reversible transformation, where the forward and the inverse mappings are computationally efficient. Here, we use the version presented in \cite{kingma2018glow}. Consider a generic split of the input vector $\mathbf{u}_j = \begin{bmatrix}\mathbf{a} \\ \mathbf{b} \end{bmatrix}$. A coupling layer is defined as: 
\begin{align*}
    & \log \mathbf{s}, \mathbf{t} = f_{\theta}(\mathbf{a}) \\
    & \mathbf{c_b} = \text{exp}(\log \mathbf{s})\odot \mathbf{b} + \mathbf{t} \\
    & \mathbf{u}_{j+1} = \text{concat}(\mathbf{a}, \mathbf{c_b})
\end{align*}
\noindent where $f_\theta (\cdot)$ is a generic transformation implemented via a NN, and $\odot$ is the element-wise multiplication. The $\log \text{det}$ of a coupling layer is simply $\sum_i \log s_i$. The main advantage of this layer is that the function $f_\theta (\cdot)$ does not have to be invertible. The main advantage is also a disadvantage, since, due to the simplicity of the affine transformation, a NF implemented with this technique needs multiple layers and blocks in order to have enough expressive power to transform any input into $p_u(\mathbf{u})$.

\textbf{2) Random Permutation}: since only a portion of the input is modified in each block, it is required to randomly permute the output of a block in order to modify a new set of parameters in the following block, i.e., $T_j(\mathbf{u}) = \mathbf{P}\mathbf{u}$, where $\mathbf{P}$ is a fixed permutation matrix. Clearly, $T^{-1}_j(\mathbf{u}_{j+1}) = \mathbf{P}^T\mathbf{u}_{j+1}$, and the transformation has a unitary determinant.

\textbf{3) Invertible Batch Norm}: As in \cite{dinh2016density}, we also apply batch normalization, but on the input of the coupling layer instead of the output. It acts as a linear re-scaling of each parameter, thus it can be easily inverted and included in the Jacobian computation. The scaling is done in the following way: 
\begin{align*}
 \mathbf{u}_{j+1} = \frac{\mathbf{u} - \bm{\mu}}{(\sqrt{\bm{\sigma}^2 + \epsilon)}}
\end{align*}
and the parameters are iteratively estimated as: 
\begin{align*}
    & \bm{\mu} = m \bm{\mu} + (1-m)\bm{\mu}_b \\
    & \bm{\sigma} = m \bm{\sigma} + (1-m)\bm{\sigma}_b 
\end{align*}
\noindent where $m \in [0, 1]$ is the momentum, while $\bm{\mu}_b$ and $\bm{\sigma}_b$ are, respectively, the mean and the standard deviation of the current mini-batch. On the first batch, the parameters are initialized as $\bm{\mu} = \bm{\mu}_b$ and $\bm{\sigma}=\bm{\sigma}_b$, and they are updated with each new batch during the training. The $\log \text{det}$ is computed as: 
\begin{align*}
    \Big( \prod (\bm{\sigma}^2 + \epsilon) \Big)^{-0.5}
\end{align*}
\noindent with $\epsilon > 0$ a parameter to avoid zero multiplication. 

\begin{figure}[t]
\vskip 0.2in
\begin{center}
\centerline{\includegraphics[width=\columnwidth]{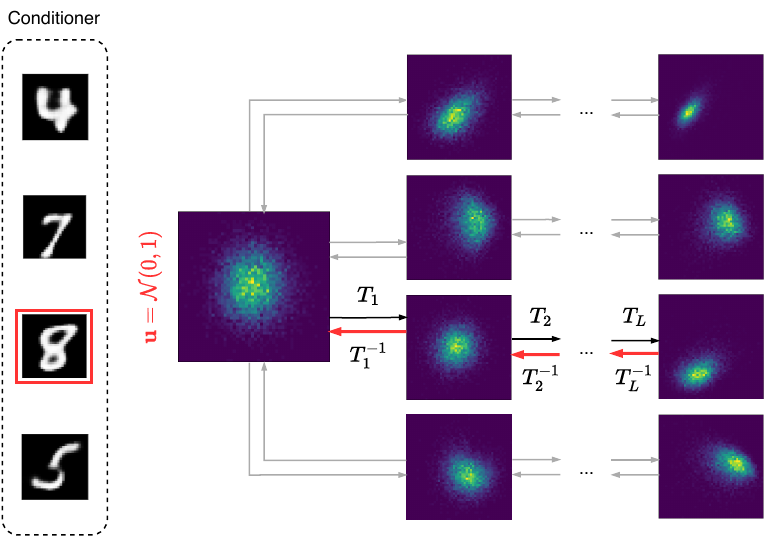}}
\caption{Visualization of the conditional embeddings produced by the trained NF on the MNIST dataset. We use PCA for visualization.}
\label{fig:embeddings}
\end{center}
\vskip -0.2in
\end{figure}

Note that, in our case, we need a NF that is conditioned on the pair $(t,y)$. To implement this, we condition the first coupling layer by passing the pair $(t, y)$ as an additional argument to $f_{\theta}$, in order to be able to generate embeddings for every possible class. The idea is similar to what is done in \cite{winkler2019learning}. Practically, the conditioning is done by building a one-hot vector with respect to the maximum number of separate classes we presume to observe. We have found the NF to be very easy to train and condition in all our experiments, with no need for specific techniques to avoid CF. In Fig. \ref{fig:embeddings} we show some examples of generated embeddings from the MNIST dataset, using several conditioning vectors.

Furthermore, we implement a multi scale architecture as explained in \cite{dinh2016density, kingma2018glow}. Each mapping $T_j$ is decomposed into multiple sub-mappings. After each sub-level, the output is split in two, with the first part being sent directly to the output of the level, while the other part flows into the next sub-mapping for further processing. This architecture improves the convergence and the stability, and it results in a smaller number of parameters.

Many others way of building a NF exists; for an in-depth review of the NF literature we refer to \cite{papamakarios2019normalizing} and \cite{Kobyzev2020NormalizingFA}. 


\section{Experiments}
\begin{table*}[t!]
\begin{center}
\caption{Average percentage on 5 runs, and the associated standard deviation, for BTW and accuracy, obtained with each method and evaluated dataset. All the results are calculated on the test sets. Best results within standard deviation are reported in \textbf{bold}.}
\label{table:results}
\vskip 0.15in
\begin{small}
\begin{sc}
\centering
\scalebox{0.9}{
\begin{tabular}{c|c|c|c|c|c|c|c|c|}
\cline{2-9}
\multirow{2}{*}{}                 & \multicolumn{2}{c|}{MNIST}                         & \multicolumn{2}{c|}{KMNIST}                        & \multicolumn{2}{c|}{SVHN}                          & \multicolumn{2}{c|}{CIFAR10} \\ \cline{2-9} 
                                  & BWT                     & Accuracy                 & BWT                     & Accuracy                 & BWT                     & Accuracy                 & BWT        & Accuracy        \\ \hline
\multicolumn{1}{|c|}{Naive}       & $-10.09_{\pm 2.46} $    & $92.95_{\pm 1.62} $      & $-5.09_{\pm 1.46} $     & $95.95_{\pm 1.04} $      & $-12.61_{\pm 1.12} $    & $84.81_{\pm 0.80}$       &    $-24.33_{\pm 5.51}$        &   $71.28_{\pm 3.66}$              \\ \hline
\multicolumn{1}{|c|}{LWF}         & $-7.17_{\pm 1.56} $     & $94.94_{\pm 1.04} $      & $-4.30_{\pm 3.64} $     & $94.44_{\pm 1.89} $      & $-12.19_{\pm 0.73} $    & $88.74_{\pm 0.48}$       &     $-24.77_{\pm 0.94}$       &         $70.71_{\pm 0.30}$        \\ \hline
\multicolumn{1}{|c|}{EWC}         & $-7.89_{\pm 1.44} $     & $94.84_{\pm 0.95} $      & $-3.60_{\pm 3.45} $     & $95.53_{\pm 2.36} $      & $-10.77_{\pm 1.10} $    & $89.32_{\pm 0.73}$       &     $-23.95_{\pm 4.12}$       &       $71.65_{\pm 4.12}$          \\ \hline
\multicolumn{1}{|c|}{GEM}         & $-0.60_{\pm 0.24} $     & $99.10_{\pm 0.16} $      & $-1.62_{\pm 0.46} $     & $97.77_{\pm 0.44} $      & $-2.41_{\pm 0.32} $     & $\bm{95.22_{\pm 0.63}}$  &        $-5.50_{\pm 0.51}$    &           $\bm{82.69_{\pm 0.69}}$      \\ \hline
\multicolumn{1}{|c|}{ER}          & $-0.36_{\pm 0.47} $     & $\bm{99.39_{\pm 0.48}} $ & $-0.73_{\pm 0.30} $     & $\bm{98.68_{\pm 0.29}} $ & $-1.63_{\pm 0.07}$      & $ \bm{95.32_{\pm 0.06}}$ &  $-4.73_{\pm 0.63}$         &           $\bm{82.22_{\pm 0.73}}$       \\ \hline
\multicolumn{1}{|c|}{PRER} & $\bm{-0.14_{\pm 0.01}}$ & $\bm{99.41_{\pm 0.10}}$  & $\bm{-0.26_{\pm 0.09}}$ & $\bm{98.70_{\pm 0.20}}$  & $\bm{-0.46_{\pm 0.17}}$ & $\bm{95.23_{\pm 0.06}}$  &     $\bm{-3.63_{\pm 0.64}}$       &        $\bm{82.67_{\pm 0.94}}$          \\ \hline
\end{tabular}%
}
\end{sc}
\end{small}
\end{center}
\vskip -0.1in
\end{table*}

\subsection{Datasets and metrics}

For evaluating the proposed method, we consider three different datasets: MNIST, KMNIST \cite{clanuwat2018deep}, SVHN \cite{Netzer2011} and CIFAR10 \cite{Krizhevsky09learningmultiple}. We evaluate the methods under the previously described multi-head CL scenario; to do so, being $C$ the classes of a dataset, these are grouped in $M$ sets, each one making up a task containing $c_m \in \mathbb{N}_+$ classes, with $c_m > 1$, giving: $M = \frac{C}{c_m}$ (
with this formulation each task contains the same number of classes, with the exception of the last one if $C\ \text{mod}\ c_m \ne 0$). 
In particular, we split the labels by grouping the original ones in an incremental way. 


To evaluate the efficiency and to compare the methods two metrics from \cite{diaz2018don} have been used. All these metrics are calculated on a matrix $R \in \mathbb{R}^{M \times M}$, where $M$ is the number of tasks, and each entry $R_{ij}$ is the test accuracy on task $j$ when the training on task $i$ is completed. This matrix of scores can be used to calculate different metrics, and we use the following:

\textbf{Accuracy}: it is the average accuracy on the trained tasks. It considers the elements of the diagonal as well as the elements below it, doing so also the evolution of the scores is taken into consideration: 
\begin{align*}
 \text{Accuracy} = \frac{\sum_{i>j}^M R_{ij}}{\frac{1}{2} M(M+1)} \,.
\end{align*}
\noindent This metric aims to show the average performance of the model in every step of the training and for each task.

\textbf{Backward Transfer (BWT)}: it measures how much information from the old tasks is remembered during the training on the new one. It is calculated as:
\begin{align*}
 \text{BWT} = \frac{\sum_{i=2}^{M} \sum_{j=1}^{i-1} (R_{ij} - R_{jj})}{\frac{1}{2} M (M-1)} \,.
\end{align*}
This metric can be greater than zero, meaning that not only the model is remembering everything about the past tasks, but it also improves the score on these. In our scenario, this phenomenon is rare, since heads associated to past tasks are no longer trained.

We use these metrics because they embed all the important aspects of the CL problem: the ability of an approach to mitigate CF and to classify correctly the past tasks, but also the ability to train the model on the current task. They are both important because a model with high accuracy and low BWT is a model not capable of alleviating CF; on the other hand, low accuracy and high BWT tells us that the constraints applied on the model are too restrictive, blocking the training of the current task.


\subsection{Experimental setting}

We compare our method to the naive approach --training on all the tasks sequentially without mitigating the CF problem-- and to the most popular or related rehearsal and regularization approaches: LWF, EWC, GEM, and ER. 

With regard to LWF, since the original method was designed for a 2-task scenario and it is not competitive outside it, we modified the approach by transforming it into a rehearsal approach: we have an external memory which contains past samples and the associated past predictions. When regularizing, the current predictions on past tasks' samples are forced to match the past prediction; in this way, this approach is competitive with the other baseline methods.

For MNIST and KMNIST we train the models using SGD with a learning rate equal to $0.1$, while for SVHN and CIFAR we use Adam \cite{kingma2014adam} with a learning rate equal to $0.001$. Regarding PRER, we train both the autoencoders and the NF using Adam, with, respectively, learning rate equal to $1\mathrm{e}{-3}$ and $1\mathrm{e}{-4}$. We use batch size equal to 64 for all datasets with the exception of CIFAR10, for which we used 128 as batch size. Each dataset have been splitted in 5 tasks by setting $c_m = 2$


We fine-tune all hyper-parameters based on a grid-search and the results from the original papers. For PRER we use 2 levels, each one composed by 10 blocks (each block is composed by: batch normalization, a coupling layer, and a random permutation); each coupling layer is defined as a fully connected network with 3 layers: $T: \mathbb{R}^{\frac{d}{2}} \rightarrow \mathbb{R}^{d} \rightarrow \mathbb{R}^{\frac{d}{2}} $. We also find that setting the coupling layers scale as $\tilde{s}=\text{sigmoid}(s + 2)$, so that the initial scale is near the identity, helps in stabilizing the training phase. 


The encoder is composed by three convolutional layers of 12, 24, and 48 filters for MNIST and KMNIST, and twice these sizes for SVHN and CIFAR10. Each layer has $4 \times 4$ kernel sizes and stride $2 \times 2$. Then, the output of the last convolutional layer is flattened and, using a fully-connected layer, the final embedding vector is produced: for MNIST and KMNIST we use 50 as embedding size, while we use 100 for SVHN and 200 for CIFAR10. The head takes as input the embedding vector, and predicts the class using three fully-connected layers, each of which halves the dimension of its input; we also use a dropout layer, with probability set to $0.2$, between consecutive layers. The activation function used in all the models is the ReLU.


We repeat each experiment for 5 times, each time changing the split of the dataset (balanced split based on labels, with proportion 80\% train and 20\% test) as well as the initial weights of the models. 



\subsection{Results}

\begin{table}[t!]
\caption{Approximate memory requirements, without taking into account the base network, and training time to do a complete training on MNIST and CIFAR10 are reported. The memory column contains, for each method, the required space (counted as number of floats to store); in order to calculate the required space we define: $N$ is the dimension of the encoder, $M$ is the number of tasks, $c_m$ is the number of classes per task, $IM$ is the dimension of an image, $E$ is the dimension of the embedding extracted by the encoder, $S$ is the number of samples saved in the external memory, and NF is the dimension of the pair (NF, decoder) used in our method.}
\vskip 0.1in
\label{table:space_time}
\resizebox{0.48\textwidth}{!}{%
\begin{tabular}{cc|c|c|c|c|}
\cline{3-6}
                            &                                 & \multicolumn{2}{c|}{MNIST}                                                 & \multicolumn{2}{c|}{CIFAR10} \\ \cline{2-6} 
\multicolumn{1}{l|}{}       & Memory                          & Memory                                                      & Time (m)     & Memory       & Time (m)      \\ \hline
\multicolumn{1}{|c|}{Naive} & -                               & -                                                           & $\approx 9$  &      -        &        $\approx 20$       \\ \hline
\multicolumn{1}{|c|}{LWF}   & $M \times S \times ( c_m + IM)$ & $\approx 1960 K $ & $\approx 20$ &       $\approx 7680 K$       &     $\approx 30$          \\ \hline
\multicolumn{1}{|c|}{EWC}   & $M \times N$                    & $  \approx 310 K            $                 & $\approx 12$ &       $\approx 2742 K$      &      $\approx 25$         \\ \hline
\multicolumn{1}{|c|}{GEM}   & $M \times S \times IM$          & $  \approx 1960 K     $    & $\approx 60$ &     $\approx 7680 K$         &     $\approx 70$          \\ \hline
\multicolumn{1}{|c|}{ER}    & $M \times S \times (E + IM)$    & $  \approx 834 K $  & $\approx 15$ &        $\approx 8180 K$      &     $\approx 40$          \\ \hline
\multicolumn{1}{|c|}{PRER} & NF $+N$                         & $\approx 148 K$                               & $\approx 20$ &         $\approx 1279 K$     &        $\approx 50$       \\ \hline
\end{tabular}%
}
\vskip -0.15in
\end{table}
In Table \ref{table:results} we summarize the results concerning accuracy and BTW, while Table \ref{table:space_time} shows the required time and memory space for each method. Results in Table \ref{table:space_time} are shown for brevity only on MNIST, but they are similar on the other two datasets. 

First, we look to the baseline methods. We can see that, in this multi-head scenario, all methods perform better that the Naive approach. LWF and EWC perform better than the naive method but not by much, and they are also more sensitive to the starting weights (as can be seen by the variance) and, in the case of LWF, also to the images stored in the external memory; furthermore, LWF performance degrades when the dataset becomes more complex, which leads us to conclude that it is not a reliable method for alleviating the CF problem. The others baseline methods, GEM and ER, performs better than LWF and EWC, with ER being slightly better in both BWT and accuracy. 
If we consider also the required time and the memory requirements, which are summarized in Table \ref{table:space_time}, we can conclude that ER is the best baseline method, because it requires less memory and less time than GEM (as advocated in \cite{pomponi2020er}).

The method we propose, PRER, achieves better BWT score than all the others methods (by a significant margin), and comparable accuracy to the best one; it is also more robust to the weights' initialization. 
Looking at the required memory space, PRER achieves the best results w.r.t. the others baseline methods, because, given a NF which is large enough, the generative model is capable of generating all the classes encountered during the training, requiring constant memory. In terms of time, it depends mostly on the dimension of the generative model, which needs to be trained separately: in the MNIST example, shown in Table \ref{table:space_time}, the required time is competitive to the other methods due to the small NF that works directly on the embeddings.

\section{Conclusion}

In this paper we introduced PRER (Pseudo-Rehearsal Embedding Regularization), a pseudo-rehearsal method that, by working on the embedding space, is able to generate past embeddings and use them to protect past information while learning new tasks. This approach is a different point of view on the pseudo-rehearsal methods, which usually work on the input space. By working on a lower complexity space, the required time is reduced as well as the dimension of the generative model. Once combined with a decoder, the generative model can be used to generate images associated to past tasks, and to constrain the model by regularizing the embeddings' deviation. We believe that this set of methods can be further investigated, leading to a different view of the pseudo-rehearsal approaches, which, right now, are feasible only for low complexity datasets. We leave an investigation on how the PRER method scales to more complex datasets to future work. 

\balance

\bibliography{references}
\bibliographystyle{icml_style/icml2020}
\end{document}